# MOSaiC: a Web-based Platform for Collaborative Medical Video Assessment and Annotation

Jean-Paul Mazellier[1,2], Antoine Boujon [1,2,*], Méline Bour-Lang[1,2,*], Maël Erharhd[1,*], Julien Waechter[1,*], Emilie Wernert[1,*], Pietro Mascagni[1, 3], Nicolas Padoy[1,2]

1. IHU Strasbourg, Strasbourg, France

2. ICube, University of Strasbourg, CNRS, Strasbourg, France

3. Fondazione Policlinico Universitario Agostino Gemelli IRCCS, Rome, Italy

* authors with equal contribution, sorted by last name alphabetical order

## Abstract

This technical report presents MOSaiC 3.6.2, a web-based collaborative platform designed for the annotation and evaluation of medical videos. MOSaiC is engineered to facilitate video-based assessment and accelerate surgical data science projects. We provide an overview of MOSaiC's key functionalities, encompassing group and video management, annotation tools, ontologies, assessment capabilities, and user administration. Finally, we briefly describe several medical data science studies where MOSaiC has been instrumental in the dataset development.

# I. Introduction

Surgical Data Science (SDS) [Maier-Hein22] represents a burgeoning research frontier that applies general Data Science methodologies to the intricate domain of medicine. In the realm of Data Science, the generation of high-quality, extensive datasets is indispensable for training algorithms, a demand accentuated in the era of Deep Learning and Big Data. While creating such datasets is already a formidable undertaking for general purposes like object detection and human pose estimation, annotating surgical datasets poses an additional layer of complexity. The sensitivity of the involved data, coupled with the imperative for clinical expertise—a resource that is both rare and costly—further complicates the annotation process.

Conventional annotation tools are typically designed for either data scientists or non-expert annotators, requiring installation on local computers and, in some instances, the compilation of source code. Moreover, their interfaces are not always user-friendly for clinicians. Additionally, the complexities of data sharing, especially in the medical field, require specific server setups and access protocols, often rendering these tools inaccessible to those outside the data scientist community.

While there is a proliferation of high-quality annotation tools—both proprietary (*e.g.* V7 labs[1] or Superannotate[2]) and open-sourced (*e.g.* CVAT[3] or LabelMe[4])—aiming to facilitate access for non-informatician communities, they remain predominantly general-purpose. Notably, these tools lack features tailored to the specific requirements of SDS, such as collaborative annotation features, medical study Group functionalities, and the seamless integration of medical ontologies. This gap in dedicated tools exacerbates the challenges faced by the SDS community in efficiently annotating medical data, hindering progress in this critical intersection of data science and surgery.

In this context, we present a new web-based annotation tool denoted as MOSaiC, which is a project derived from a previous collaborative work [Mazellier23]. With respect to the initial project, significant efforts have been devoted to proposing a seamless tool for collaborative and guided work on medical video annotation/assessment tasks. We focus on a Group-centric approach providing a complete and coherent framework for physicians and data scientists to jointly work on the creation of medical video datasets. This tool is already actively used by about 250 users working on 3000 hours of procedures spread over some thirty projects.

# II. Why and how did we develop MOSaiC platform?

Dealing with dataset collection and annotation for more than 10 years, IHU Strasbourg[5] and the CAMMA research Group[6] have built strong expertise in this domain. This process of annotation is a time-consuming effort where user retention is critical. We identified different key elements that can strongly affect motivation over time, among which:

- In the specialized domain of SDS, data annotation and curation efforts require not only data scientists but also the valuable involvement of clinical partners. While data scientists rely on the dataset as an essential component for their daily work, which revolves around data analysis, clinical partners typically undertake this task in addition to their primary clinical priorities.

---

[1] https://www.v7labs.com/
[2] https://www.superannotate.com/
[3] https://www.cvat.ai/
[4] https://labelme.csail.mit.edu/
[5] https://www.ihu-strasbourg.eu/en/
[6] https://camma.u-strasbg.fr/

- Complicated annotation tools with steep learning curves can be demotivating for clinical partners. A complex, secondary task might be deprioritized if it puts a strain on clinical priorities.
- Each single medical data science study has its particularities, nevertheless, designing an annotation tool that is specialized for each single study is intractable.
- Auditing data quality is central to annotation to ensure the development of effective algorithms.

Beyond the data science perspective, extensive discussions with clinicians revealed a notable absence of an efficient tool tailored to their needs for video assessment. For instance, assessment scales such as OSATS (Objective Structured Assessment of Technical Skill) are increasingly used in surgery as tools for measuring surgical quality and safety, enabling objective evaluation of performances [Dzau21]. These tools are generally applied to video-based assessment using non-specific software like spreadsheets. This is not only inefficient but also error prone. Since clinical video-based assessment is fundamentally an annotation task as intended in the data science domain, we chose to propose a dual tool allowing to perform both tasks in one platform: assessment and annotation.

From these key points, we defined some basic principles that stand at the core of MOSaiC development. We can list these main principles as:

- **Keep it simple but not simplistic:** the medical annotation process is complex by nature; nevertheless, we pay particular attention to offering the simplest tools, interface, mechanisms, etc. while not constraining ourselves to only simplistic approaches. This balance is always complicated to ensure, nevertheless, it demonstrated to be highly efficient in promoting user adhesion to the tool. One essential point to figure out is the best way to proceed via constant communication with different medical experts to define their "perceived" needs, in order to translate them into software development requirements. This continuous close relationship allows us to find a convenient tradeoff for an efficient platform both for clinical and data science users.
- **Keep it guided but not rigid:** one complex task is to avoid over-specializing the platform for a few studies with features that won't be reused easily in other studies. We dedicated significant effort to collecting needs for the different studies to propose tools and mechanisms that could be tuned to every need in the simplest way. Doing so, every single study can be set from the general tools in the proper way. We avoided to propose a common, single, rigid framework for all studies that would have limited user retention. As a consequence, we do not enforce a general structure that every study must follow, but we offer a strong foundation of general-purpose annotation functionality that could be refined with minimal configuration work to adapt to the particularities of every single study.
- **Keep communication central but do not overwhelm users:** users need to communicate efficiently to perform their tasks, this is a general paradigm that everyone can witness with the current development of collaborative platforms. Medical data science does not represent an exception here. We took extra care to offer a centralized communication mechanism helping users to communicate with each other, for follow-up purposes, to annotate collaboratively, and to stay updated, etc. On the other hand, we minimize exposure to extra information as much as possible to avoid defocusing users who work on cognitively demanding tasks.

At the heart of MOSaiC developments are these fundamental yet often overlooked principles. While seemingly basic, they form the foundation of our approach. This is why we decided to develop our annotation tool, MOSaiC, building on previous work [Mazellier23]. In the rest of the document, we will detail all the features we developed and explain how the platform can, even if not completely but significantly, cover the needs we identified in our work of dataset creation for medical data science.

Before proceeding, we expose here the two main profiles we can identify for MOSaiC users:

- **User**: any person with a validated account in the MOSaiC platform
- **Annotator**: a user dedicated to proceeding with the annotation task as required by a study. This role is crucial for the dataset creation. To ease the tasks of Annotators, the platform can be tuned to provide only essential tools, timely updates, and clear communication processes.
- **Group Manager**: a user able to fine-tune MOSaiC to the study requirements, both in terms of other user profiles as well as in terms of available tools, review mechanisms, etc. This role is central to any study. Note that Group Manager possesses all permissions in the Group (see Section III.A. Management of Annotators in a Group).

Note that these profiles are not strict in the sense that they are not enforced on the platforms. On the contrary, we chose to leave complete tuning possibilities for defining study-specific "profiles" to ensure the general approach described earlier. In particular, the same user can be Group Manager of Group "A", being Annotator in Group "B" and do not have access to Group "C".

# III. Description of the MOSaiC annotation tool[7]

In this Chapter, we first introduce the notion of MOSaiC Group, a central concept in all the development and use of the platform. These Groups are designed to provide a relevant framework for the work in annotation that is described in Paragraph IV. We then describe the different features of the tool.

## A. Group

MOSaiC is developed with a Group-centric approach to help and guide Annotators in producing high-quality datasets. This structure allows us to natively silo data and users in the most relevant way: users can only access relevant material they are allowed to, and they are guided to focus on the tasks they are assigned to.

Each user's first view of MOSaiC lists all the Groups this user has access to (Figure 1). Each Group's basic features are:

- o Group's set of videos (1) to be annotated.
- o Video bookmarks (2) for each user to quickly reach the videos of interest.
- o Group's ontology (3).
- o Group's document section, easing access to study-specific documentation like annotation protocols (4).

---

[7] Version 3.6.2

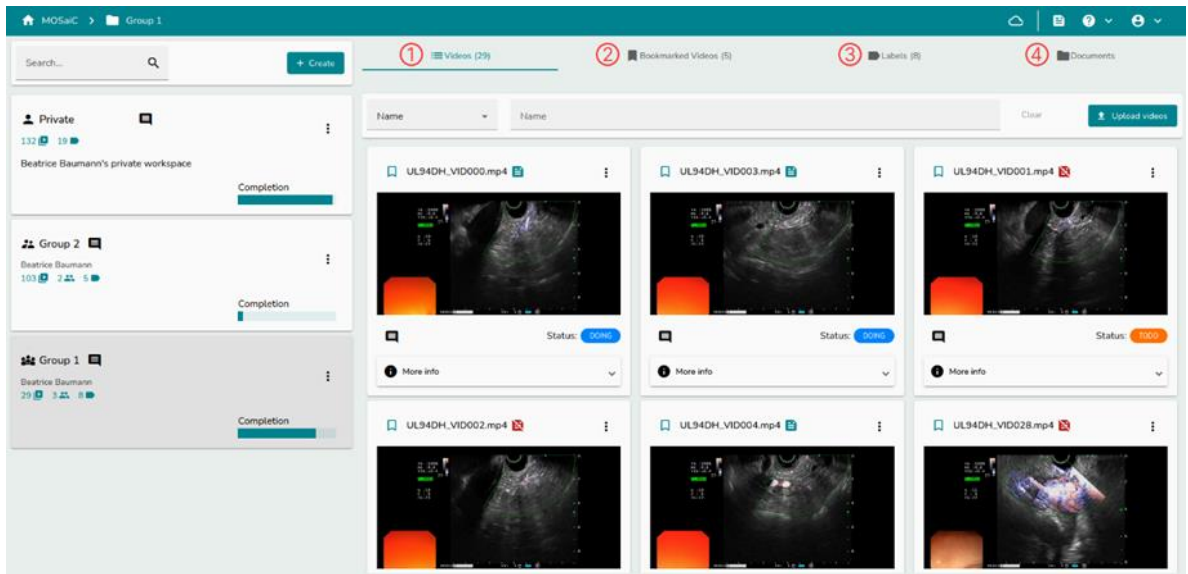

*Figure 1. Groups view as presented to a particular user. Only Groups that the user has been assigned to (by Group Managers) are visible.*

There are 3 types of Groups:

- Collaborative: Annotators work together on the videos with unrestricted interactions in between each other. Simultaneous work is possible, all annotations being updated in real-time and visible by all Annotators in the Group.
- Supervised: Annotators have access to a video once it has been specifically assigned to them by Group Manager, the annotations are not shared between Annotators (i.e., Annotators are blinded to each other annotations), and the guidance can be adapted to the study through the different features as described later.
- Private: By default, each MOSaiC user is assigned a Private Group with Group Manager status. This Group contains all the videos uploaded by the user and provides a "playground" to annotate and keep work private.

When creating a Group, the Group Manager can choose its type (collaborative or supervised) and create or import an existing ontology from another pre-existing Group. Selecting a Group shows its content (videos and ontology labels). In the Private Group, this allows the users to manage their videos (rename, assign a clinical protocol – see Section III.C.e -, or delete the video). In any Group, Annotators can filter videos (by name, label used, or annotation status), bookmark them, update their annotation status, define data characteristics, or see their statistics in the Group context (labels used, Annotators, etc.).

**Which Group type should I use: collaborative or supervised?**

This choice is mainly driven by the flexibility/guidance level the Group Manager wants to offer to Group's Annotators.

Collaborative Groups allow for efficient multi-user co-annotation tasks [Fleurentin22, Graeff23, Kassem22, Lavanchy22, Lavanchy23, Meyer22, Yu23]. Multiple Annotators working on the same video can see current activity, in particular, if another Annotator is working simultaneously on the same material. Discussion threads are available at different levels: per Group, per video, and per annotation. This can improve communication among potential Annotators who are physically spread across different locations. Annotators can create any annotation and modify or delete other Annotators' annotations (if they have the permissions to do so, see Section III.A.e). A history is available to cancel last actions in case of unwanted changes.

Supervised Groups share a large portion of collaborative Group features while offering dedicated possibilities among which:

- Video access (Figure 2):
    o Annotators are assigned a subset (up to the full set) of videos in the Group by the Group Manager.
    o Each video is assigned a level. This level can for instance be associated with the difficulty of the recorded procedure.
    o Each Annotator is given an initial access level. This level can be changed afterward.
    o An Annotator can only access videos that are both assigned and of equal or lesser level compared to the Annotator level.

*Figure 2. Video assignment page in a supervised Group. In this view, a Group Manager can view each Annotator's level, assigned and hidden videos. Videos can be assigned/unassigned to Annotators through this interface. Note that each video has a level. Only assigned and compatible level videos (i.e. videos with lower or equal levels to the Annotator) are effectively visible to the Annotator. For instance, on this screenshot, in "Assigned Videos" panel, the two upper videos are not accessible (grey shaded) to current Annotator because their level is 2 while Annotator's level is 1.*

- Individual access:
    o Annotators can only access and modify their work, making the annotation process blind to other Annotators. This way Annotator's assessments are independent, a particularly important feature for when assessing Annotators inter-rater variability is needed.
    o Group Managers can have access to every Annotator's work. It is possible to export all the annotations (*i.e.* from every single Annotator) of a Group at once.
- Automated evaluation:
    o Predefined annotations can be granted as ground truth by the Group Manager. By doing so, once an Annotator completes assigned videos annotation, a score can be calculated based on a comparison of the produced annotation with respect to the predefined ground truth (see Figure 3). This allows an automated level-up mechanism, providing access to the next level of videos.

*Figure 3. Example of a form (that can be associated with any type of annotation). In this particular case, a ground truth has been defined by a Group Manager and Annotator-submitted answers are evaluated automatically.*

- o The ground truth can only be configured by the Group Manager, using answers of a specific Annotator (including Group Manager their self).
- o All Annotators can see their progression in the Group by visualizing the percentage of videos set as "DONE" compared to level compatible accessible videos.
- o Once all the assigned videos are set as "DONE" and the completion reaches 100%, Annotator answers are compared to the ground truth to calculate a score. If the score exceeds a preset threshold, the level of the Annotator automatically increases to allow them to work on more difficult videos.

The supervised Groups demonstrated efficiency in blindfold multi-Annotator dataset creation [Graeff23, Vannucci22] and also when annotation upscaling (in particular regarding the number of involved Annotators) is required as in the ongoing SAGES CVS Challenge[8]. This will be detailed in the Section Examples of use cases in MOSaiC.

### a. Group management

There are several views dedicated to Group management:

- o Add or remove videos from the Group. Only the user (with "Video uploader" role, see Section III.C.Administrator application) who uploaded a video on the platform can share it in a Group. The same video can be shared in multiple Groups.
- o Add or remove users from the Group and manage their permissions.
- o Manage a supervised Group: assign or hide videos from Annotator, change an Annotator's level, and set a particular Annotator's annotations as the ground truth in the Group.
- o See Group statistics: graphs of the time spent on videos (per Annotator); time spent on Group per day (per Annotator).
- Uploading a video is always done in the context of a Group (either in the Private Group or directly in the target Group). When the users upload one or multiple videos, they can choose their video level and protocol. Once started, the progress of the upload will be displayed in the main toolbar.
- It is possible to export all the annotations of a Group to a JSON file.

---

[8] https://www.cvschallenge.org/

   b. Management of videos in a Group

In order to ensure good practices for medical data access, a clinical protocol is attached to each video. This protocol indicates the framework in which data has been acquired and the goal of the study, gathers all administrative information (*e.g.* IRB number) and the deadline for data archiving, easing the quality control of the data management process. For public data, dedicated protocols may be defined allowing flexible use of the data. Once a video is attached to a protocol, it can be shared with users through multiple Groups for dedicated annotation tasks.

   c. Management of Ontology in a Group

In MOSaiC, the ontology is a collection of labels (Figure 4), the last one being of one of two types:

- <u>Temporal</u>: annotations will be defined by a timestamp on a start frame and a duration specified by marking the timestamp of the end frame or by explicitly providing the number of frames associated with the annotation.
- <u>Spatiotemporal</u>: bounding box, point, polygon, polyline, or segment. The related annotations will have the same temporal properties and include the coordinates of the shape over time.

Each label has a name, a color, and a type.

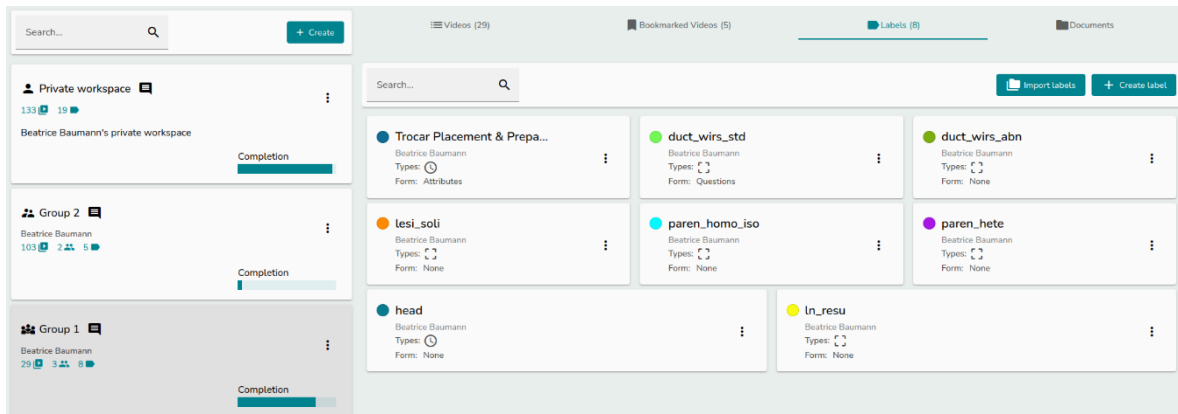

*Figure 4. Ontology page view. For each Group, a specific set of labels constitutes the Group's ontology. Each label has specific features (name, type, and color).*

Note that for handling cases of complex ontologies, a label-Grouping feature has been introduced that allows multilevel tree sorting of annotation labels. This tree configuration can be easily changed without any impact on previously created annotation. This is particularly useful for complex phase-steps surgical video annotation [Lavanchy22, Lavanchy23] and it also helps preserve a compact timeline for better clarity (Figure 5).

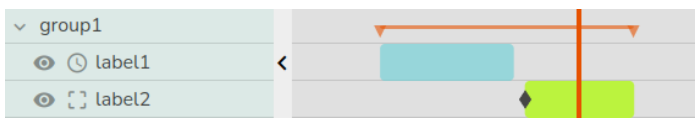

*Figure 5. Ontology elements can be Grouped in a multilevel tree hierarchy in order to ease the process of visualization in the annotation timeline.*

In addition to standard features (color, name, type), each label can optionally be attached with a form, which is a way to add information about a (spatio-)temporal annotation on top of its basic features (like timestamps and coordinate if spatial). There are two possible types of forms:

- Attributes: All Annotators answer in the same mutually shared form (particularly one Annotator can modify another's answers).
- Questions: Annotators with corresponding permissions can create an annotation with a question form. Such annotation is available to any Annotators (even the ones with no right to create annotations themselves). If an Annotator is granted permission to answer forms, this Annotator is allowed to individually populate the annotation form blinded from other Annotators. This allows to create rubrics useful for video-based assessment.

Creating or editing the form opens a dedicated form-building tool with a preview (Figure 6). Currently, 22 types of questions are available (true/false, choice between preset numbers, custom select menu, free text…).

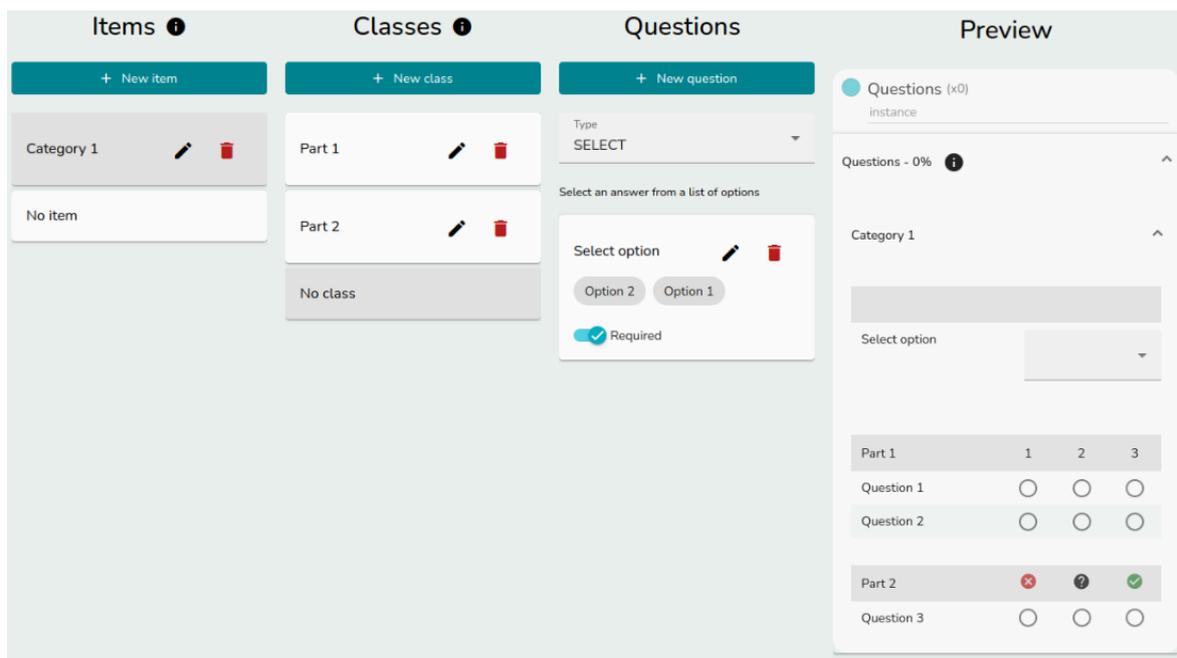

*Figure 6. Form conceptor. In order to ease features/questions Grouping, 2 levels have been introduced as (in order from top level to bottom level) "Items", "Classes", "Questions".*

Note that the labels of a Group can easily be imported into the ontology of another Group to speed up the process of creating a new Group with similar (or close) ontology. In this case, modifying the ontology in one of these two Groups will not affect the ontology in the other.

### d. Management of information in a Group

Each Group provides by default a page for sharing information among every Annotators in the Group. Currently, only PDF documents can be stored and shared. In particular, this space is convenient for sharing an up-to-date annotation protocol or any other information that any Annotator of the Group should access frequently. This simple yet efficient feature ensures a relevant Grouping of information, so Annotators do not lose time searching for a specific point in their annotation work, increasing their engagement in the long term.

### e. Management of Annotators in a Group

Access to a Group can be granted by Group Manager to any MOSaiC registered user. Once integrated, the user is considered an Annotator of this Group. Individual Annotators profiles settings can then be defined by the Group Manager to improve guidance and restrict possible actions. For instance, a "basic" Annotator will only have permission to create annotations in videos, but no possibility to modify the ontology, add/suppress videos, etc. The Group Manager automatically has all permissions in the Group.

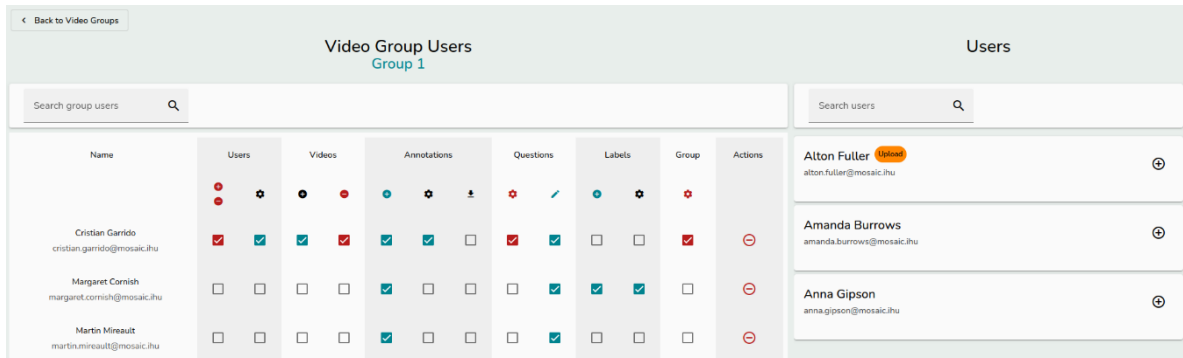

*Figure 7. Group's Annotators management page allows to onboard users in the Group as well as set their permissions in this particular Group.*

The different possible permissions are described in Table 1.

*Table 1. List of possible Annotator's permissions*

| Permission | Description |
| --- | --- |
| add/remove users | Add/remove users from the Group and assign them any Group permission |
| manage user access | Change Annotator's level, video level and annotation state, in order to be able to assign videos to Annotators. Has access to all videos in the Group. |
| add videos | Add videos in the Group. The videos can come from the user or from another Group |
| remove videos | Remove videos from the Group. This does not delete the videos, but it deletes all data associated to the Group (annotations, form answers, …) |
| create annotations | Create spatial or temporal annotations (and attributes) and edit/delete them. To create annotations with questions, permission "Manage questions" is also required. |
| manage annotations | Edit/delete annotations (and attributes) created by other users. To manage annotations with questions, permission "Manage questions" is also required. |
| download annotations | Download all annotations of the Group (without restrictions on level or annotation state) |
| manage questions | With "Create annotations", create annotations with questions; with "Manage annotations", edit/delete annotations with questions created by other Annotators (if an annotation with questions is deleted, all the associated answers are deleted as well). With "Create labels", create labels with questions; with "Manage labels", build questions forms for labels. |
| answer questions | Answer annotations with questions. |
| create labels | Create annotation labels (with optional attributes). To create labels with questions, permission "Manage questions" is also required. |
| manage labels | Edit/delete annotation labels (and attributes). To manage labels with questions, permission "Manage questions" is also required. |
| edit Group | Edit the Group: edit name and description, manage documents. |

## B. Annotation page

This interface is the core part for the Annotators. It allows producing the annotation on a particular video in the way the Group Manager organized the process. An example of an annotation page is presented in Figure 8.

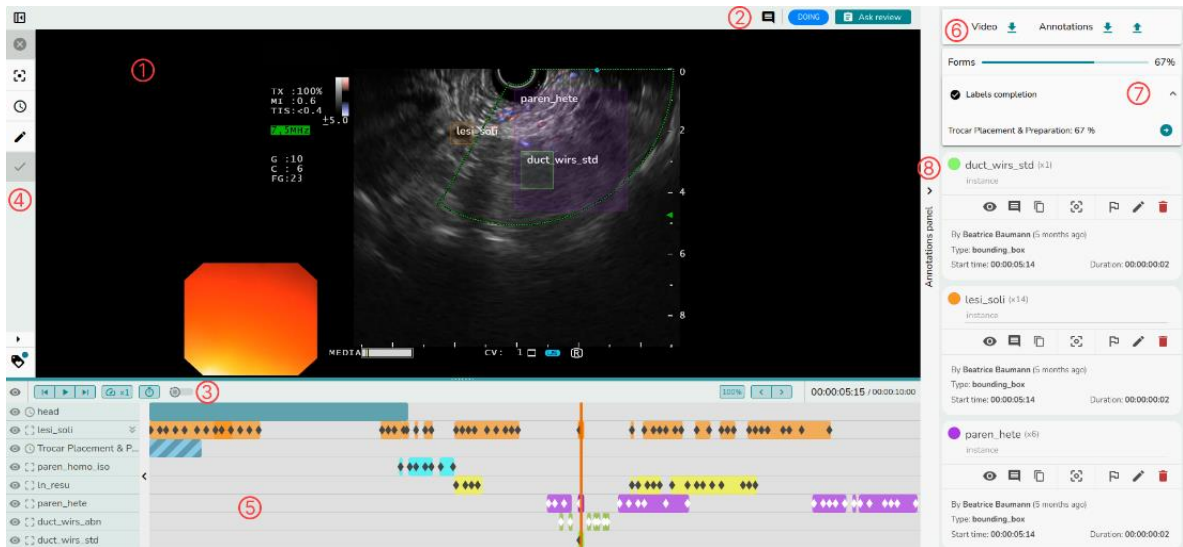

*Figure 8. Screenshot of the annotation page with its different panels described in the section.*

- The annotation page of MOSaiC is divided into different sections:
    o <u>Video player (1):</u> displays the video and the spatial annotations with their labels. Depending on the Annotator's settings, the annotation positions can be interpolated (using linear interpolation). The video player height can be resized until the timelines are hidden.
    o <u>Top toolbar (2):</u> allows to manage the video annotation state or get to the next video to annotate. In a collaborative context, it also lists the current participants.
    o <u>Video toolbar (3):</u> allows to manage the video (play/pause, go to start/end of the video, change speed (x0.2 to x5), jump to previous/next annotation, change displayed time format).
    o <u>Annotation toolbar (4):</u> allows to create or edit annotations. It can be hidden.
    o <u>Timelines (5):</u> contain the temporal information of all the annotations of the video, Grouped by label (and ordered by start frame or label name). It displays which frames contain spatial annotations and shows when an annotation has comments or an incomplete form associated with it. It is possible to zoom in on the timelines and get to a frame-level precision. The timelines also allow the hiding of specific labels/annotations, unfold overlapping annotations of the same label, and automatically Group labels by prefix.
    o <u>Import/export (6):</u> allows to download the video or export all the annotations of the video in a JSON format and to import annotations into the video using the same file format.
    o <u>Form progression (7):</u> if the annotations of the video are associated with forms, display the global progress, and progress for each label. This also allows Annotators to jump to the next incomplete form.
    o <u>Annotations panel (8):</u> lists all the annotations corresponding to the frame currently displayed in the video player. The Annotator can edit or delete the annotations, fill out

their forms, and comment on them. An annotation can also be duplicated. This panel can be hidden.

A set of keyboard shortcuts exists to make the work of the Annotators easier; a detailed list is directly available in the application. This includes a basic "undo" feature.

### a. Annotation creation/edition

- <u>Temporal annotation creation:</u> when an Annotator creates a new annotation, they choose a label (or create a new one). After this step, the annotation automatically begins at the current frame of the video. The Annotator can then move in the video (either by clicking in the timelines or playing the video) and validate the annotation when it reaches its last frame. Temporal annotations of one-frame can easily be created in one click thanks to a dedicated button.
- <u>Spatial annotation creation:</u> after the label has been selected, the Annotator will need to draw a shape on the video to save the current frame as the annotation's first frame. The type of shape (bounding box, point, …) is defined by the chosen label. The shape can be edited at any following frame of the video: its new coordinates will be saved into the annotation. Note that for the spatio-temporal annotation, MOSaiC makes use of keyframe interpolation to ease the annotation process. Annotators can define particular features of the spatial annotation for some frames (keyframes) with MOSaiC automatically calculating these features for each frame in between two consecutive keyframes. For bounding boxes, points, and segments, MOSaiC uses linear interpolation of attributes. For polygons and polylines, we replicated the interpolation algorithm as implemented in CVAT annotation tool[9].
- <u>Annotation edition:</u> the temporal information of an annotation (start frame, number of frames) can be edited in the annotations panel, or directly in the timelines. An annotation can also be cut into multiple ones; if it has a shape, the positions will be interpolated to keep the cut annotations consistent with the old one. The shapes of a spatial annotation can be edited directly on the video player, as it is done at its creation. An instance field can be used to track the same object cross multiple annotations.

Hereafter (Figures Figure 9 and Figure 10), we present some spatial annotations screenshots from the MOSaiC platform.

---

[9] https://www.cvat.ai/

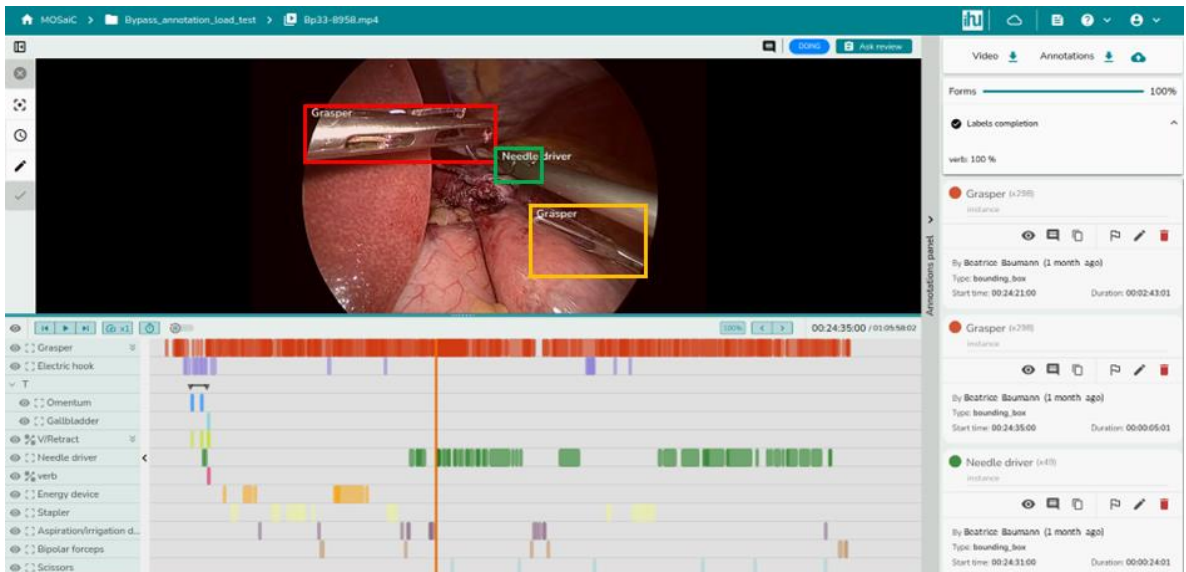
*Figure 9. Example of bounding box annotation in the MOSaiC platform.*

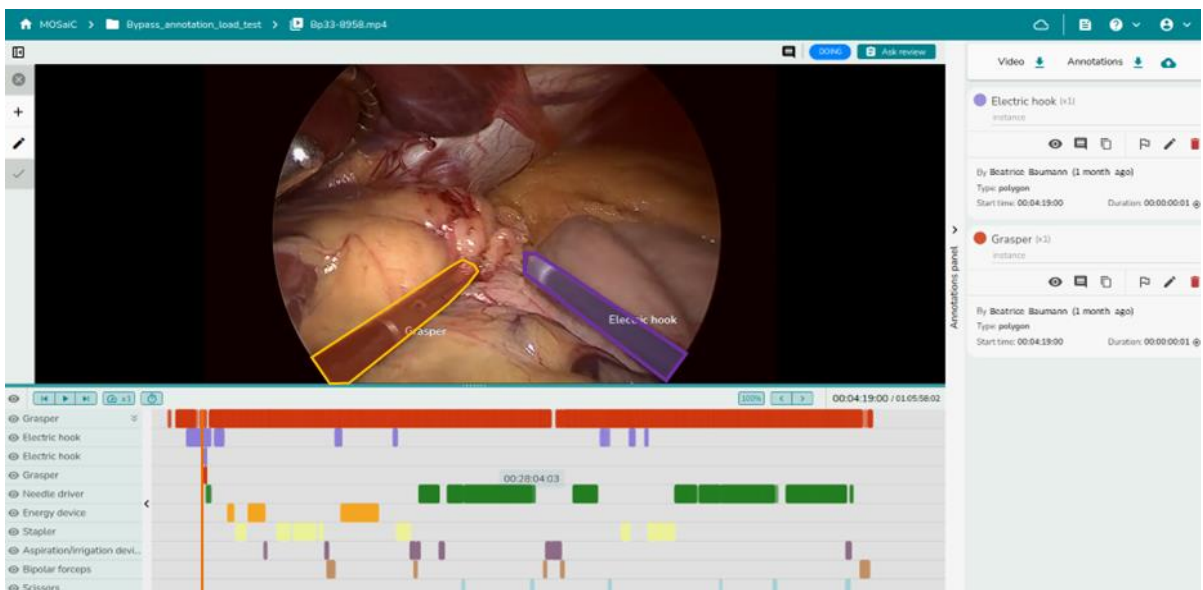
*Figure 10. Example of segmentation annotation in the MOSaiC platform.*

### b. Video status

In order to keep track of the annotation state for every single video in a Group, a status is attached to the videos. This status can be changed by Annotators to inform about the action to be performed. MOSaiC offers (see Figure 11):

- NEW: video has been added to the Group but no annotation has been performed yet (default status).
- TO DO: indicates that is video has to be annotated in priority.
- DOING: indicates that the annotation work is under progress (default status once the first annotation is added to the video).
- REVIEWING: Annotators can ask for a review of their annotation work to pre-defined reviewers.

- DONE: indicates reviewers have validated the Annotator's work. Note that reviewers can change the status to "DOING" and ask for specific revisions to Annotators if needed. This cycle takes place until the reviewer validates the video to "DONE" status.

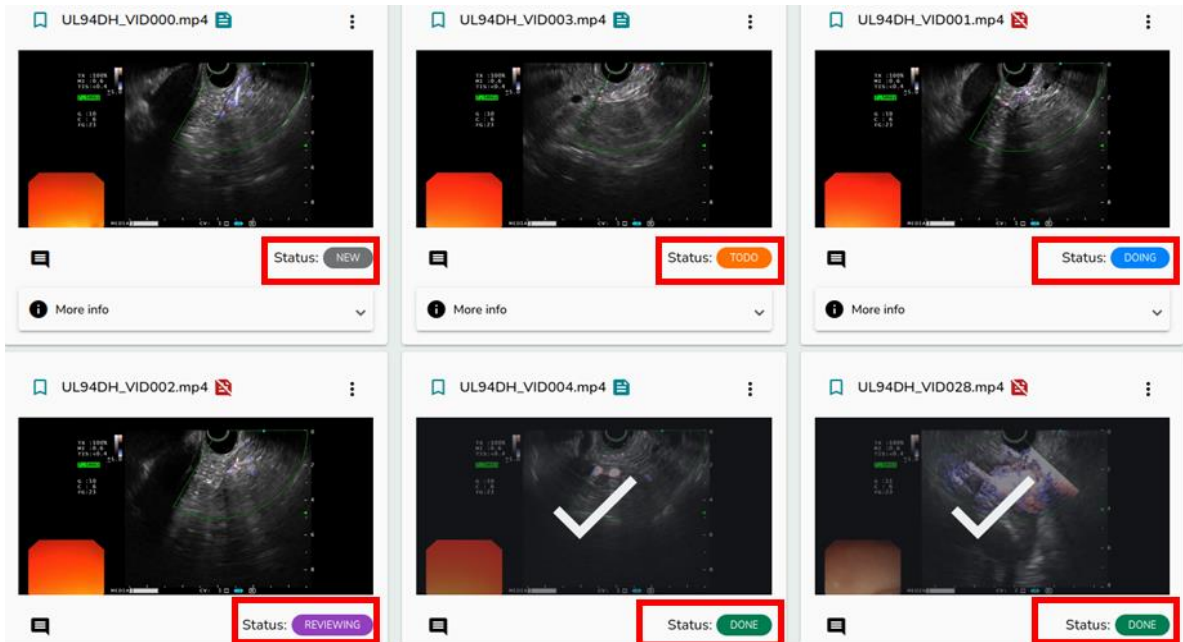

*Figure 11. Different video thumbnails and their associated status (bottom-right of each video panel). This status can be viewed in the Group view (Figure 1) and changed in the Annotation view (Figure 8).*

### C. General Features

General features are by default present in MOSaiC, easing the overall process of annotation and easing communication between users. Hereafter, we detail some of these features: two factor authentication to the platform, adaptative bitrate streaming, comments, dashboards, and protocols.

#### a. Landing page and Signup/Login views

MOSaiC web landing page allows users to enter the platform. This landing page is presented in Figure 12.

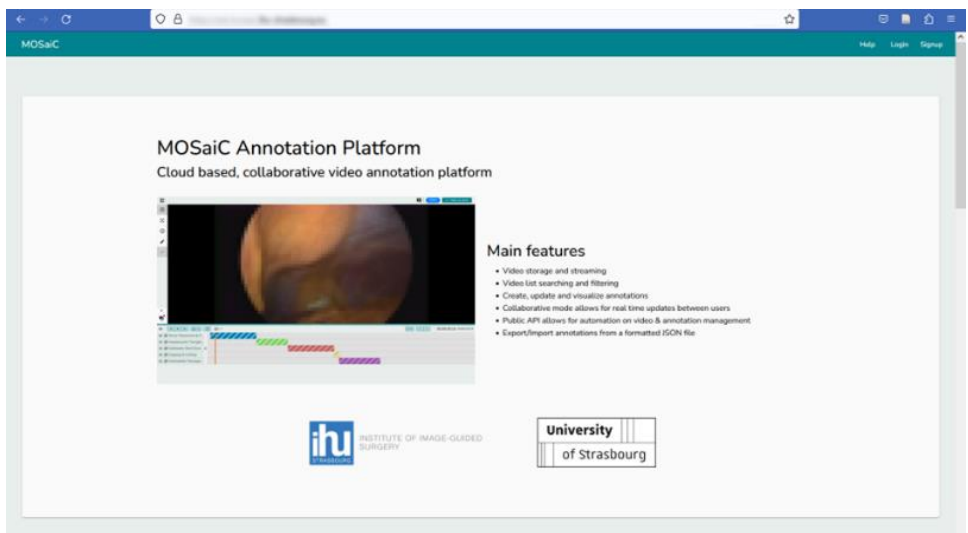

*Figure 12. MOSaiC platform landing page. In the upper-right corner, there are links to "signup", "login" and redirect to relevant documentation.*

Dealing with medical data, even if pseudonymized, requires particular care surrounding system security. With MOSaiC, the overall security is reinforced with the following features:

- Access to MOSaiC requires users to create an account protected by a password (Figure 13.a). Users can request an account using a signup form. To log in, they will need to wait for manual activation by administrators after security checks (known user, valid project, etc.). We point out that, even after this activation, users have by default no access to any data. This data access is conducted for each case through the control of a Group Manager (Section III.A.Management of Annotators in a Group).
- After activation, any access is secured via a two-factor authentication mechanism (Figure 13.b). The user will be required to enter a code received by email (as associated with the user account) to log in.

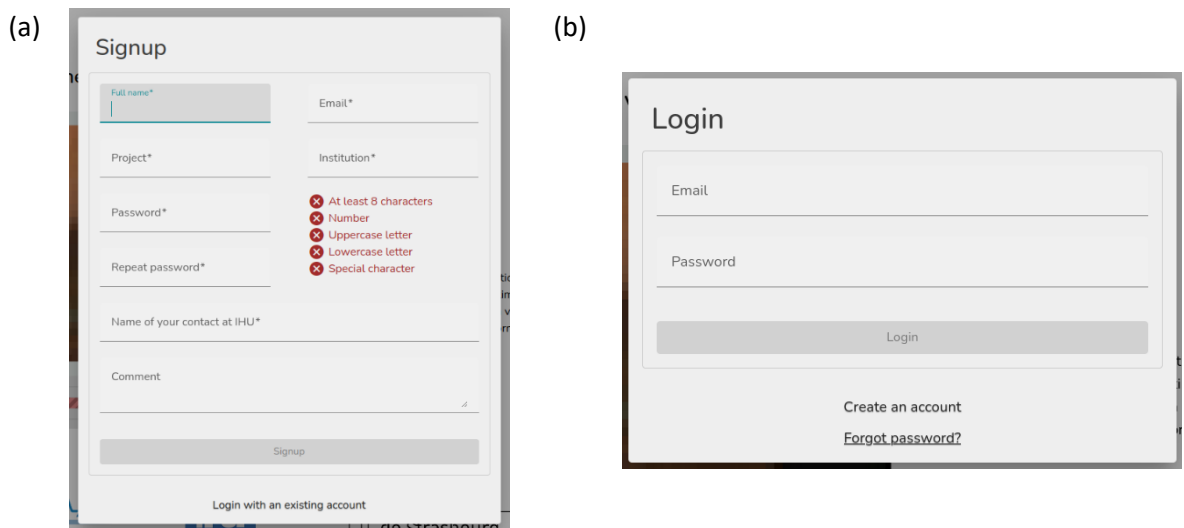

*Figure 13. (a) Signup page. An account creation is the first step to enter the MOSaiC platform. Credentials (login and password) and additional information (concerned project, contact, institution) are requested to ensure complete identification of any new user; (b) Login page. Once an account is validated by MOSaiC platform administrators, a user can enter the platform through the login page where credentials are asked. If valid, a two-factor authentication is performed by sending a code to the user on the associated account email. The user has to enter the code to enter the MOSaiC platform.*

b. Adaptative Bitrate Streaming

MOSaiC makes use of Adaptive Bitrate Streaming (ABR) for video playback when annotating. This feature is particularly useful for Annotators struggling with low bandwidth or an unstable network. Every time a user uploads a video through MOSaiC, the video is re-encoded and split into segments, but the original video still exists in storage. After this process, the video is available in different resolutions: 4k, 1080p, 720p, 480p, 360p, 240p and 144p. The maximum resolution available will depend on the original video resolution. Any video with a resolution lower than 144p will not be processed and the video player will stream the file directly. The video can then be played through a protocol named HTTP Live Streaming (HLS). The Annotator's interface uses an open-source video player named *HLS.js*[10]. It automatically adapts the displayed video resolutions according to the

---

[10] https://github.com/video-dev/hls.js/

Annotator's bandwidth. Since the video is divided into segments, the resolution switch operates seamlessly during the video playback.

Note that the fact that MOSaiC is video-oriented is central to the treatment of temporal (and spatio-temporal) annotations. In particular, web video plugins are predominantly time-based. However, the notion of frame is essential in the subsequent data science analysis. We chose to expose a frame-based interface to users to ensure consistency in dataset generation with future dataset usage. This conversion relies on the calculation of the frame index (for analysis) from time (as imposed by web video players). If obvious for most videos, complex video encoding can make this translation risky, because of rounding errors that could be browser-dependent, or when dealing with variable frame rate videos. We strongly encourage users to re-encode any potential problematic video to enforce at least constant framerate. Under this constraint, we can ensure the explicit use of videos instead of image-based approaches (like with CVAT for instance), enhancing the possibilities in data streaming on the web platform.

### c. Comments in MOSaiC

Comments threads can be added to different entities in MOSaiC (Groups, videos, annotations) to help communication between users. In a supervised setting, Annotators can only interact with Group Managers. A comment thread can be resolved for closure. If an annotation has comments, this will be displayed in both the timelines and the annotation panel (see Figure 14).

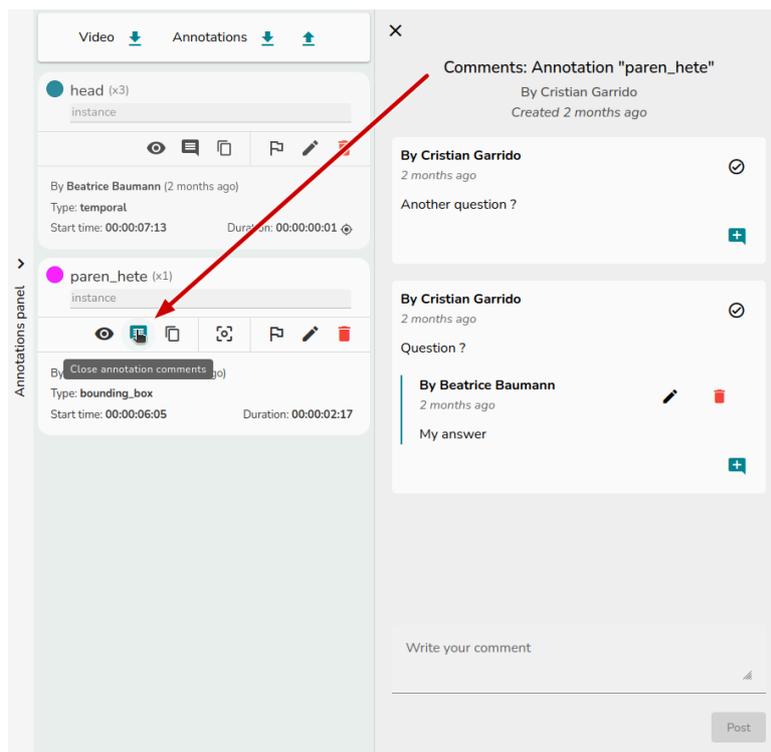

*Figure 14. Example of a comment thread as displayed in the annotation panel.*

### d. Activity tracking

In order to help with the annotation auditing, different dashboards are available to the Group owner. In particular, a per video activity dashboard indicates how long each Annotator in the Group has spent working on a particular video (Figure 15). A similar dashboard, but per Annotator (Figure 16), indicates activity in the Group over time. This eases the process of annotation by pointing out potentially complex videos or difficulties in the annotation work for a particular Annotator.

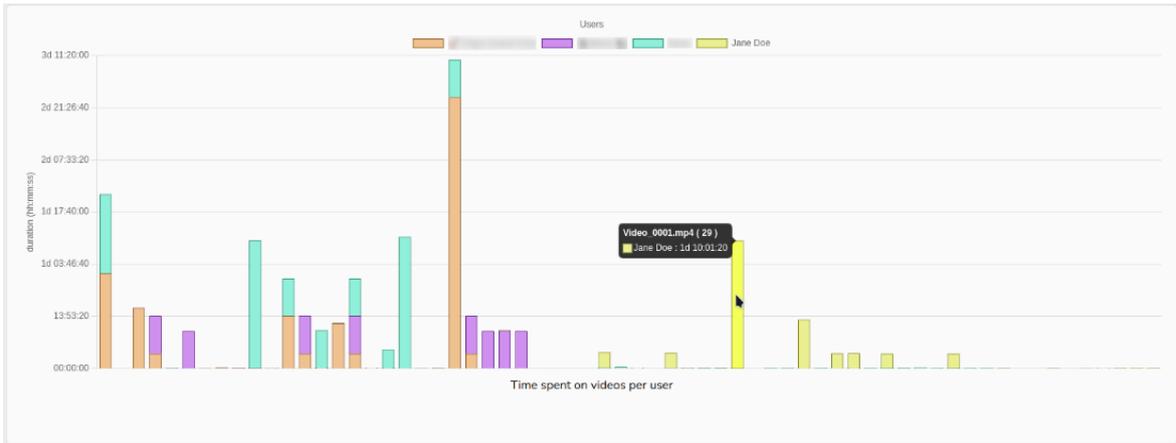

*Figure 15. Per video time spent dashboard (specific to a Group). The time each Annotator spent on videos is reported in cumulative histogram format.*

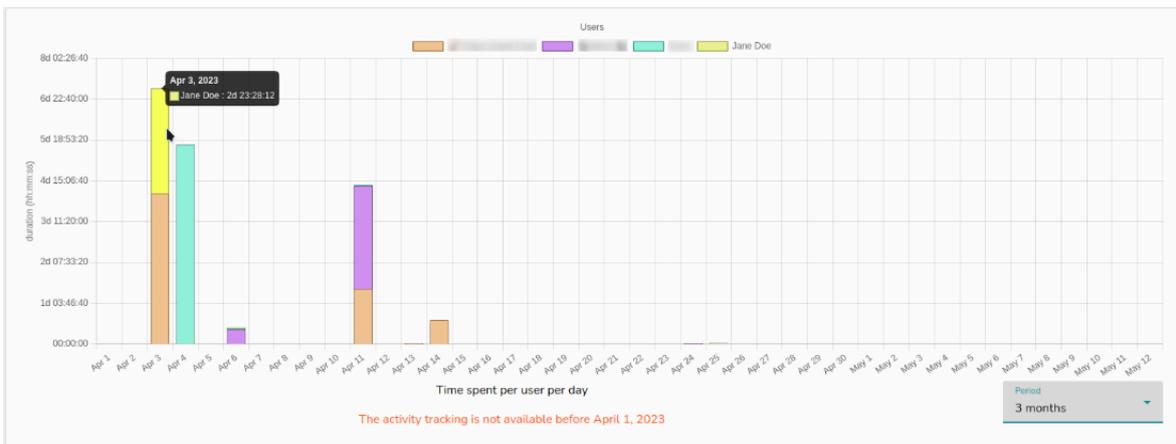

*Figure 16. Monitoring of historic of activity (in a specific Group) dashboard. A per day view of each Annotator's activity is provided.*

### e. Medical study protocols

Video collection and processing often have to be validated by an institutional review board or ethical committee. In MOSaiC we provide a way to attach such information to every single video to ensure data control and conformity with best practices. In particular, protocols specify what is the time and extent of data use before archiving. When creating a protocol, all relevant information can be filled in to ease the process of medical data management in the long run. A view is dedicated to the management of protocols (Figure 17). Protocols are files that can be associated with videos to describe the experiment or study protocol (how the videos were acquired and how to use them). Only uploaders that were granted access to a protocol can assign it to their videos.

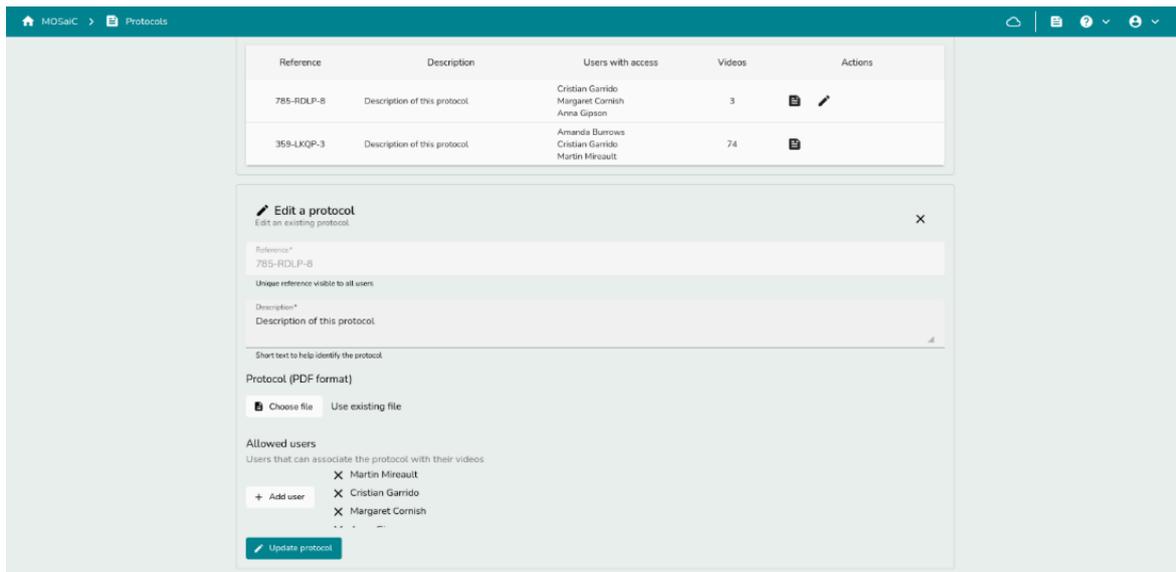

*Figure 17. Protocol view enabling creation of a protocol entry with associated information. This protocol can then be attached to different videos related to it.*

## f. Administrator application

MOSaiC has a distinct application dedicated to administrator users (Figure 18). It mainly allows to:

- See users, activate/disable/archive them, define their roles
- See active connections to MOSaiC
- Set global settings (activate/deactivate features, 2-factor authentication, basic customization...)
- See all videos on the stack, rename or delete them
- Define terms of use that should be accepted by users on MOSaiC
- Send mail to all users

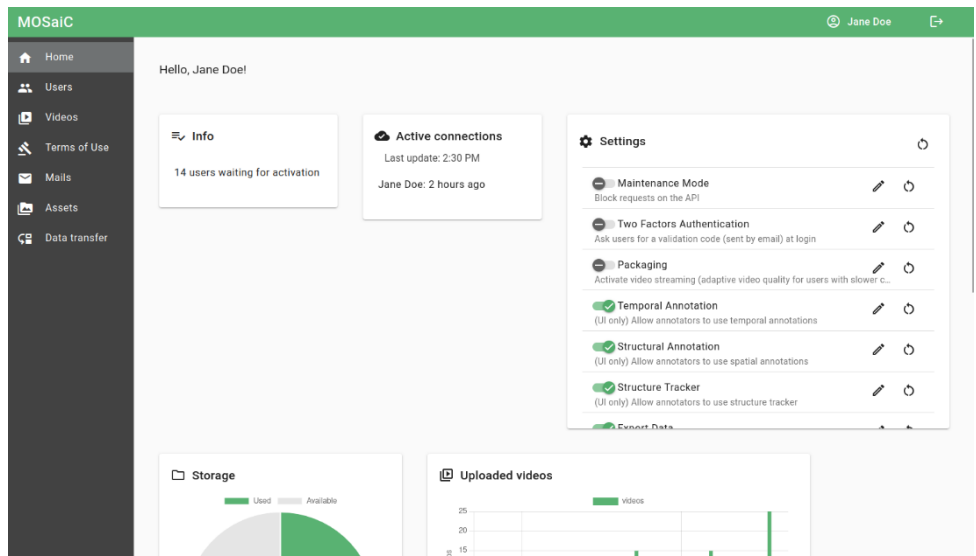

*Figure 18. MOSaiC administration webpage.*

**Roles**

In the MOSaiC administration application, users have roles (given by administrators), that can be combined:

- Administrator: can do anything on the platform and has access to all Groups.
- Group creator: can create video Groups and share them with any user on the platform.
- Video uploader: can upload videos on the platform and owns a private workspace (= Group containing all videos uploaded by the user).
- Protocol manager: can see all protocols, manage them (create/edit/delete), and allow uploaders to associate them with their videos.
- Script user: can create API tokens, valid for a predefined duration, which can be useful to run scripts.

# IV. Examples of use cases in MOSaiC

As stated in the introduction, MOSaiC has been developed to cover unmet needs and provide the flexible structure required to reach the various requirements encountered in the multiple medical data science studies. Here, we will describe some of the studies that have used MOSaiC as an annotation platform for the dataset creation as well as video assessment.

## A. Collaborative annotation

The most straightforward way to proceed with dataset production is to provide one or more users with the videos to be annotated in one common interface, provide an annotation protocol to do so, and let the Annotator(s) perform the task. This scenario can be enabled by using a MOSaiC collaborative Group. In this Group, videos, ontology, and instructions are centralized and easy to access. The Annotators can focus on performing their task, with MOSaiC natively hiding irrelevant functionalities and providing quick access to the annotation protocol if Annotators have doubts. This helps keep the Annotators focused on their complex primary task - annotation. Once finished, Annotators have the opportunity to ask for a review by changing the status to "Reviewing". Reviewers are notified and can check the annotation. Here two possibilities can be encountered:

- Annotations conform to the expectation and fit the annotation protocol, as judged by reviewers. The video can be validated, and the video status is changed to "DONE".
- Some part of the annotation has to be corrected. Here reviewers have access to specific labels: if ontology possesses a label "A" (accessible to Annotators), reviewers have access to a label "correct_A" of the same type as "A". This helps reviewers point out a problem with an annotation but also gives Annotators clear feedback on what should be corrected and how. For instance, an incorrect bounding box can be pointed out, and the reviewer can draw a correct bounding box in a part of the video to help Annotators modify their work accordingly. Once reviewed this way, the reviewer changes the status of the video to "Doing" and Annotators are notified. They can modify their annotations under the guidance of the "correct_" labels and validate each requested correction. They can then ask for a second review. This process is carried out and iterated until it is validated by reviewers.

This scenario is the one that most annotation platform offers. It is conceptually simple and offers great flexibility. However, MOSaiC distinguishes itself on many points in the general organization of the annotation task:

- The overall uncluttered ergonomics of the platform (Group view, accessible documentation, clean annotation view, etc.) is a strong point, validated and refined through the different feedbacks of clinicians. The web interface is almost a must-have feature, enabling easy access to any Annotators with no local installation requirement but access to a general-purpose web browser.
- This multi-actor annotation process requires the strong involvement of the Group Manager to ensure constant quality of the annotation, needing to generate motivation and maintain engagement from the Annotators. For this, the communication tools offered by MOSaiC, such as comments and notifications, proved themselves of high importance.
- The reviewing features collection (video status, false positive flagging, correction proposals, annotation history, etc.) has demonstrated its strength through the iteration-refinement process with a cycling Annotator pool.

This pattern has been implemented for instance in the following different projects:

- The ABITO project [Yu23] for temporal phases annotation over 80 videos from the Cholec80 dataset [Twinanda17]
- The APEUS-AI project [Fleurentin22, Meyer22] for temporal phases and bounding box annotations on almost 200 videos so far, involving 8 annotators located in France, Italy and Germany
- The FedCy study [Kassem22] for temporal phase annotation over 180 laparoscopic cholecystectomy videos collected from 5 european medical centers, involving 3 annotators in Italy and France.
- The Gastric By-Pass study [Lavanchy22, Lavanchy23] for complex temporal phases/steps on almost 70 videos. For this last study, the MOSaiC collaborative Group feature allowed efficient annotation between four annotators spread among different countries (France, Switzerland, and Italy) in a seamless way, despite the very complex ontology of surgical phases/steps involved (see Figure 19).

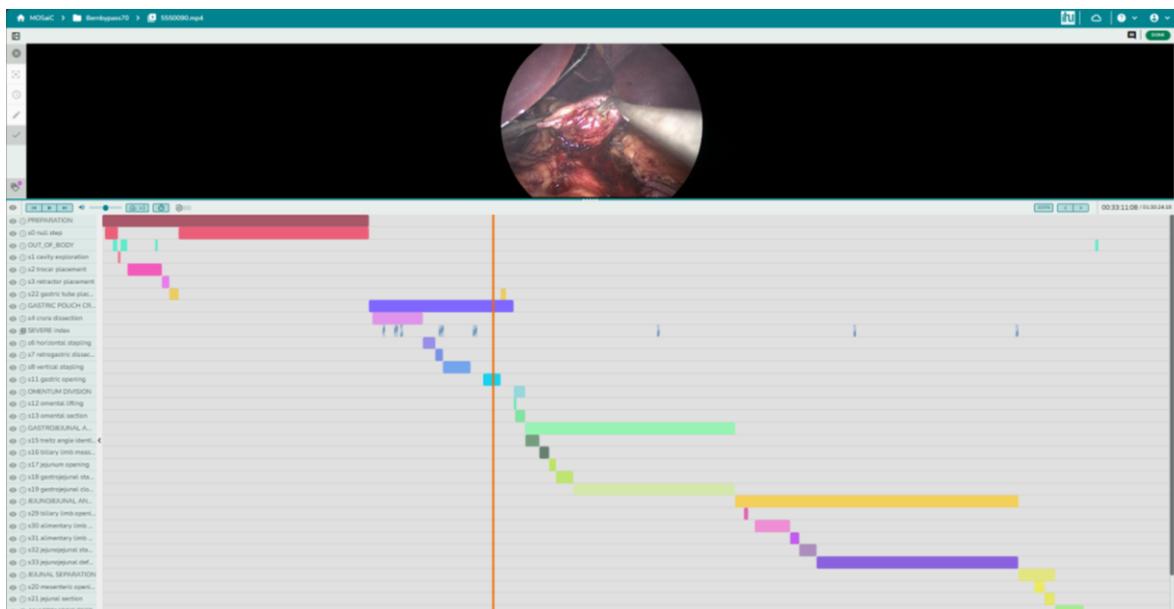

*Figure 19. Example of a complex surgical phase annotation for a gastric by-pass procedure. This annotation is part of a collaborative Group that allowed direct interaction within a Group of four Annotators. The created dataset has been used in [Lavanchy22, Lavanchy23] for automated surgical phase/step recognition.*

### B. Guided video-based assessment

MOSaiC is also being used to facilitate collaborative and highly granular video-based assessments. For instance, in the context of a project related to assessing laparoscopic cholecystectomy operative difficulty [Vannucci22], 3 surgeons were asked to assess more than 70 items impacting the complexity of procedures (e.g., severity of adhesions, anatomical variations, etc.) and scales assessing surgical performances (e.g., OSATS, SEVERE, etc.) on 100 laparoscopic cholecystectomies videos. Initially, surgeons were sharing videos using cloud-based storage and performing the video-based assessment on predefined excels, with each spreadsheets listing items to be assessed for each phase. We transferred this complex organization into MOSaiC by creating a supervised Group. In this Group, the Group Manager annotated phases and associated to each phase a form presenting all phase-specific items to be annotated. Each Annotator was then able to visualize phases and empty forms or rubrics prompting surgeons to independently assess phase specifics items informing on laparoscopic cholecystectomy operative difficulty and surgical performance. The burden of video-based assessments was significantly decreased thanks to MOSaiC reducing the time spent annotating, making the process less error prone and overall increasing users' engagement. Here MOSaiC allowed to propose a perfect tradeoff between guidance and study organization, allowing Annotators to keep focused on their surgical assessments, removing all unnecessary constraints for data access or annotation (see annotation interface on Figure 20).

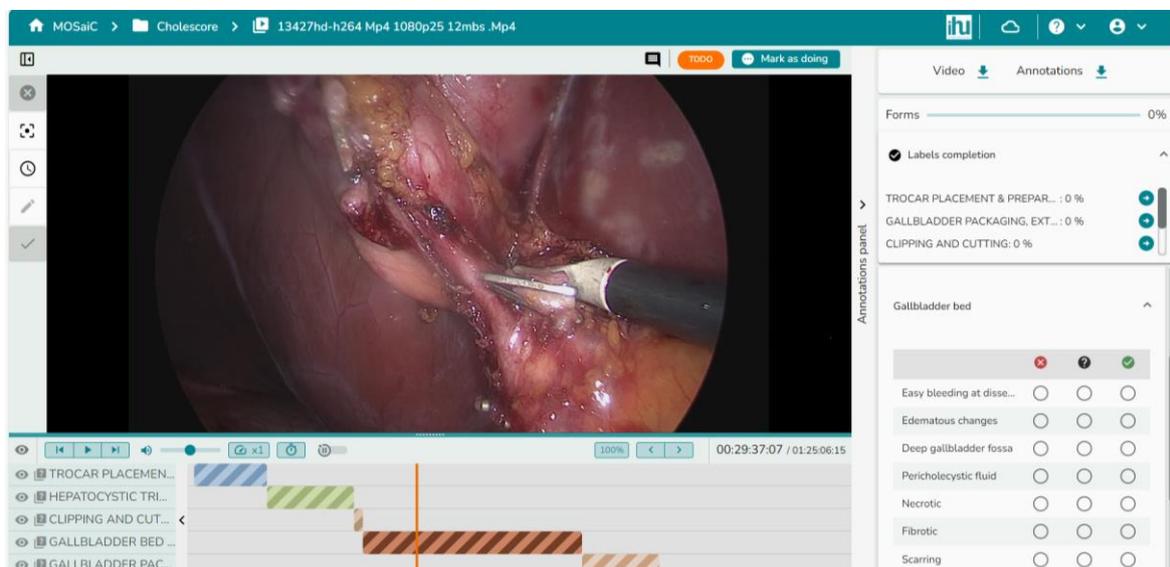

*Figure 20. Laparoscopic cholecystectomy difficulty assessment form as implemented on the MOSaiC platform. Specific temporal annotations have been defined by Group Manager for the different phases. For each phase, a dedicated form is available for Annotator to provide guided assessment via different features (here time cursor is on the "Gallbladder bed dissection" phase, with associated form displayed in the annotation panel on the right). The dashed aspect of the temporal annotation indicates incomplete form. Also, all form completeness can be assessed with progress bar (on top of the annotation panel) with completeness progression percentage and direct access to unfilled forms (right arrows below the from progress bar). All these features help Annotator to have a quick overview on potential missing information before submitting answers. Note: submission is possible only if required fields (as defined in the ontology) are filled.*

### C. Annotation at scale: the "SAGES CVS Challenge"

In the SAGES CVS Challenge[11], the goal is to produce a sizeable and diverse dataset of laparoscopic cholecystectomy videos and related annotations on the achievement of the Critical View of Safety (CVS) [Mascagni21a, Mascagni21b] by:

- Collecting an unprecedented number of de-identified laparoscopic cholecystectomy videos from medical centers all over the world.
- Screening all videos according to the study eligibility criteria

---
[11] https://www.cvschallenge.org/

- Editing videos to the region of interest for CVS assessment
- Assessment of the 3 criteria defining CVS in each video clip and on evenly spaced frames.

The eligibility assessment of videos is performed in MOSaiC in a dedicated Group (see Figure 21), where de- identified videos are uploaded from a central collection server thanks to a developed API (see Section V). Once uploaded, each video is labeled with a timestamp signaling the region of interest and with punctual information on eligibility annotated through a form. An automated scripts checks the annotations on eligibility criteria and either edit or exclude the video. If the procedure is deemed eligible, a short clip is extracted and uploaded to a second Group dedicated to CVS annotation. During the upload, the pre-defined forms prompting to assess the achievement of CVS criteria at the case level (video clip) or on evenly spaced frames are automatically added. Finally, clips and forms are made available to Annotators.

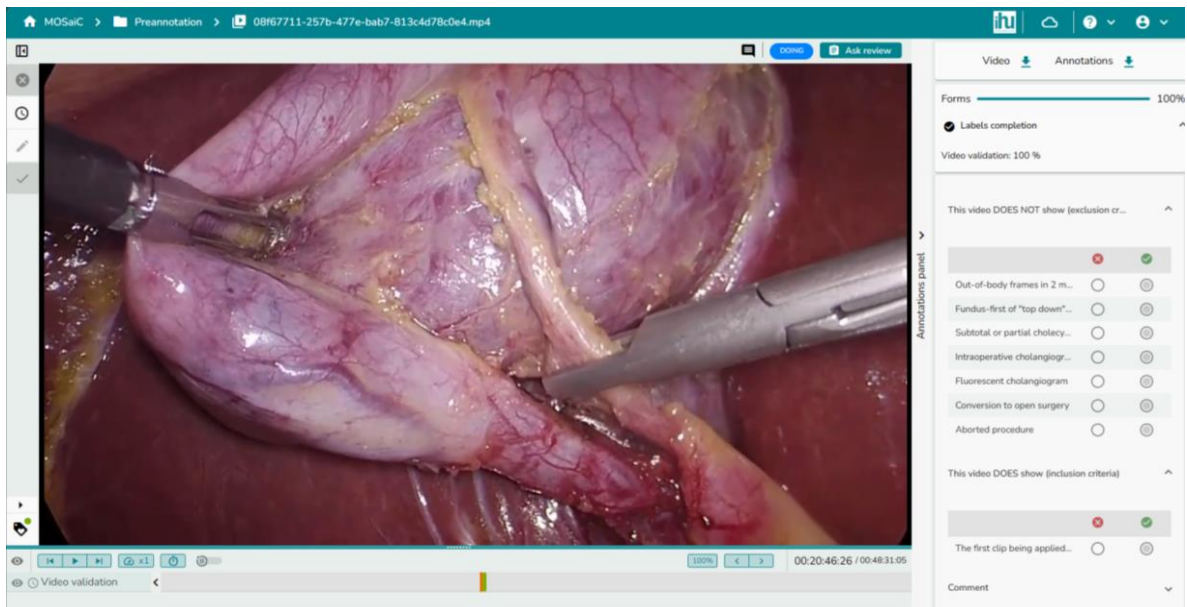

*Figure 21. Video eligibility assessment annotated using pre-defined criteria. Annotators are asked to annotate the first surgical clipping action in the video, with a one-frame annotation, and then populate the associated form where all the criteria are listed as binary options (yes/no). Based on the created annotation, videos can be filtered out if ineligible, or clip extraction is performed based on the annotation timestamp. Eligible clips are then made available for annotation in MOSaiC.*

In addition, a annotation school was designed using a dedicated supervised Group to ensure data quality throughout the SAGES CVS Challenge (see Figure 22). Experts defined solid ground truth annotations for video clips included in the school. Annotator candidates are granted access to the supervised annotation school Group (with annotation protocol available in the "Documents" panel) and are asked to assess included videos. Once done, their answers are compared to experts' consensus ground truth and if an accuracy of at least 75% is reached, they are onboarded in the annotation part of the challenge. This annotation school ensures that users reach a minimum quality of annotation (e.g. by following a predefined annotation protocol as defined by metrics of inter-rater agreements) before being able to independently annotate videos (see Figure 23).

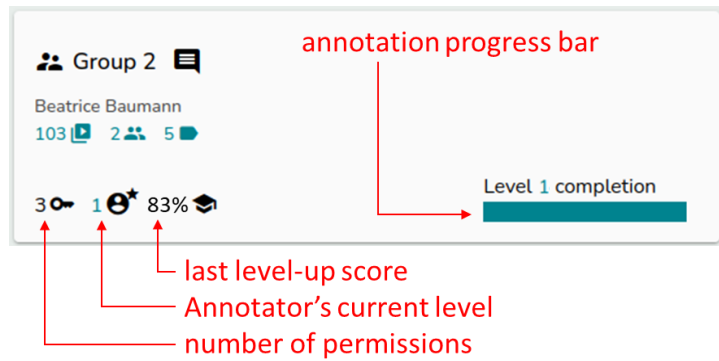

*Figure 22. Individual Annotator progress in the annotation school Group (based on a supervised Group). Here the annotator has completed with a score of 83% the 1st (and unique) level of difficulty, giving access to autonomous annotation phase. The Group Manager has access to individual Annotator's progress and each assigned video's current status, easing follow-up efforts. The Annotator's level-up mechanism is automated in MOSaiC as soon as a ground truth and metrics (here accuracy) are defined.*

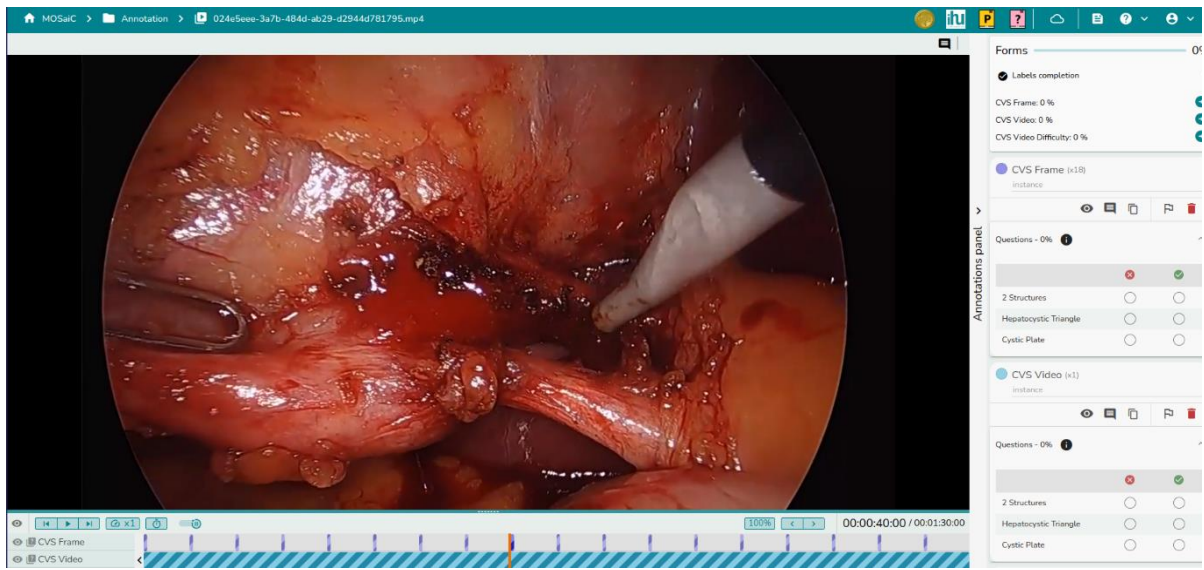

*Figure 23. Annotator's view available during the autonomous annotation phase once annotation school has been passed. The time cursor is located on a one-frame temporal annotation for frame-wise CVS assessment (CVS frame on annotation panel). A video level CVS assessment is also available (CVS video on annotation panel). In both cases, the three CVS criteria can be assessed via YES/NO radio button elements.*

The SAGES CVS-Challenge[12] required a new level of scale where MOSaiC proved to be an efficient way to cover the required automatic training curriculums, blinded annotations (because CVS assessment is prone to inter-rater variability), review processes due to high consequence of human error to flag Personal Identifying Information, API access for scripts to build a complex system that MOSaiC had to be integrated into (video collection/anonymization pipeline, cloud storage, etc.). During the whole process, MOSaiC proved itself an essential tool to scale the annotation process for selecting eligible videos out of thousands of files. The annotation school is also a strong possibility in MOSaiC, helping in managing tens of candidates before onboarding for the annotation phase of the study. Then MOSaiC offers adequate guidance in this core annotation phase, allowing Annotators to focus on their task seamlessly, not being disturbed by external factors.

---

[12] https://www.cvschallenge.org/

## V. MOSaiC architecture and dependencies

MOSaiC 3.6.2 is a web-based platform to ease access for multiple users, especially clinicians, seamlessly. It is conceived to be cloud native and can be deployed on on-premises servers as well as on most general cloud provider platforms thanks to containerization. It is based on the following frameworks:

- API: NestJS 10.2
- UI: Angular 16.2
- Admin App: Angular 16.0
- Structure Tracker: NestJS 9.3
- Worker: 10.0

And uses:

- Docker
- Kubernetes
- Postgres: 15.4
- Redis: 7.2

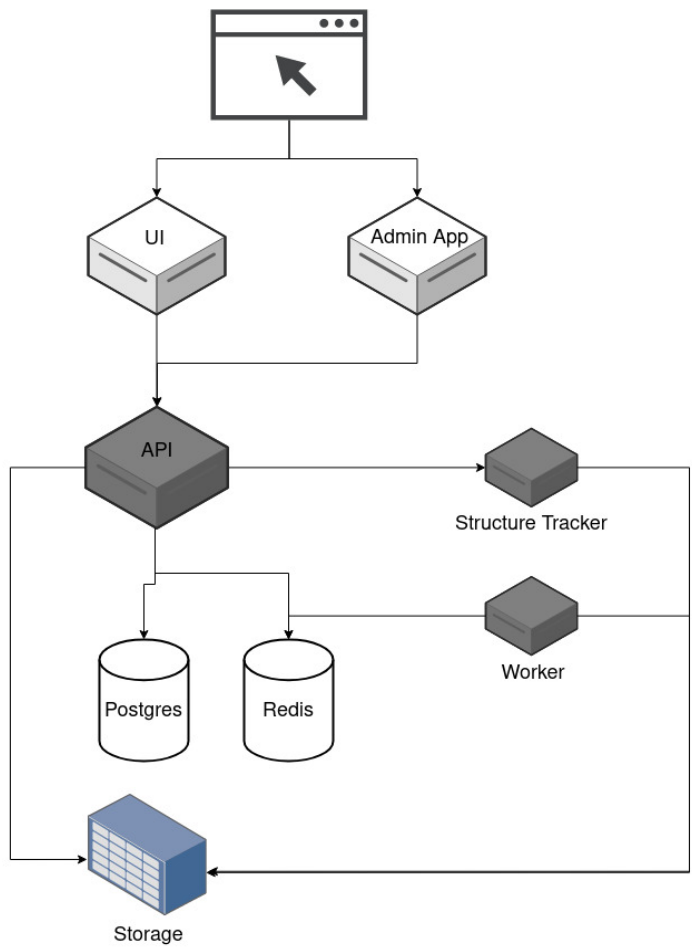

*Figure 24. General architecture of the MOSaiC platform.*

# VI. Conclusion

MOSaiC is a web-based annotation tool developed to propose a specific platform dedicated both to clinical studies and to the creation of medical datasets for computer vision tasks. The broad spectrum of potential users, ranging from medical students and experts to data scientists, requires a combination of collaborative and quality management features that makes MOSaiC a unique platform. In particular, extra care has been devoted to proposing a simple, yet comprehensive interface where only the tools needed are available to annotators to provide the best user experience and reinforce engagement in annotation tasks. This has improved the adoption of the tool by the medical community and reduced the errors in the dataset creation, ensuring high-quality results for better exploitation through machine learning algorithms. MOSaiC proposes a Group-centric approach, making it possible to tune the tool and its features per project and make the deployment of annotation tasks as easy as possible. Collaboration and review are central concepts in MOSaiC that make annotation production more efficient even with teams spread across the world. In addition, the supervised Groups allow us to easily create annotation schools to ensure Annotators' conformity with predefined protocols, and also to evaluate inter-rater agreement when producing datasets with multiple annotators. The increasing number of users and video material demonstrate the value of MOSaiC as a tool that can help produce high-quality surgical data science datasets at scale.

# VII. Acknowledgements

This work was supported by French state funds managed within the Investment for the Future Program by the ANR under references ANR-10-IAHU-02 (IHU Strasbourg), ANR-22-FAI1-0001 (DAIOR) and by BPI France under reference DOS0180017/00 (project 5G-OR). It was also supported by the ARC Foundation (www.fondation-arc.org) within the APEUS project.

[Maier-Hein22] L. Maier-Hein et al, Surgical data science – from concepts toward clinical translation, Medical Image Analysis, 2022

[Mascagni21a] P. Mascagni, D. Alapatt, A. Garcia, N. Okamoto, A. Vardazaryan, G. Costamagna, B. Dallemagne, N. Padoy, Surgical data science for safe cholecystectomy: a protocol for segmentation of hepatocystic anatomy and assessment of the critical view of safety, arXiv abs/2106.10916, 2021

[Mascagni21b] P. Mascagni, A. Vardazaryan, D. Alapatt, T. Urade, T. Emre, C. Fiorillo, P. Pessaux, D. Mutter, J. Marescaux, G. Costamagna, B. Dallemagne, N. Padoy, Artificial Intelligence for Surgical Safety: Automatic Assessment of the Critical View of Safety in Laparoscopic Cholecystectomy using Deep Learning, Annals of Surgery, 2021

[Mazellier23] J.-P. Mazellier, M. Bour-Lang, S. Bourouis, J. Moreau, A. Muzuri, O. Schweitzer, A. Vatsaev, J. Waechter, E. Wernert, F. Woelffel, A. Hostettler, N. Padoy, F. Bridault, INDEXITY: a web-based collaborative tool for medical video annotation, ArXiv abs/2306.14780, 2023

[Meyer22] A. Meyer, A. Fleurentin, J. Montanelli, J.-P. Mazellier, L. Swanstrom, B. Gallix, G. Exarchakis, L. Sosa Valencia, N. Padoy, Spatio-Temporal Model for EUS video Detection of Pancreatic Anatomy Structures, Proceedings of MICCAI Workshop on Simplifying Medical Ultrasound, 2022

[Nwoye23] C.I. Nwoye, T. Yu, S. Sharma, […], N. Padoy, CholecTriplet2022: Show me a tool and tell me the triplet — An endoscopic vision challenge for surgical action triplet detection, Medical Image Analysis, 2023

[Twinanda17] A.P. Twinanda, S. Shehata, D. Mutter, J. Marescaux, M. de Mathelin, N. Padoy, EndoNet: A Deep Architecture for Recognition Tasks on Laparoscopic Videos, IEEE Transactions on Medical Imaging (TMI), arXiv abs/1602.03012, 2017

[Vannucci21] M. Vannucci, G. Laracca, P. Mercantini, S. Perretta, N. Padoy, B. Dallemagne, P. Mascagni, Statistical models to preoperatively predict operative difficulty in laparoscopic cholecystectomy: a systematic review, Surgery, 2021

[Yu23] T. Yu, P. Mascagni, J. Verde, J. Marescaux, D. Mutter, N. Padoy, Live Laparoscopic video Retrieval with Compressed Uncertainty, Elsevier Medical Image Analysis, 2023
26